\documentclass[letterpaper,11pt]{article}

\usepackage[margin = 1in]{geometry}
\usepackage[utf8]{inputenc}
\usepackage{amsmath}
\usepackage{graphicx}
\usepackage{acronym}
\usepackage{authblk}
\usepackage[colorlinks=true,allcolors=blue]{hyperref}
\usepackage[backend=biber,style=numeric-comp,sorting=none,safeinputenc=true]{biblatex}
\AtEveryBibitem{
 \clearfield{eprint}
 \clearfield{isbn}
 \clearfield{issn}
 \ifentrytype{inproceedings}{}{
 \ifentrytype{misc}{}{
  \clearfield{url}
  \clearfield{urldate}
  \clearfield{urlyear}
 }
 }
}

\begin{document}

\title{Capturing reduced-order quantum many-body dynamics out of equilibrium via neural ordinary differential equations}

\author{Patrick Egenlauf$^{1,}$\footnote{patrick.egenlauf@simtech.uni-stuttgart.de} , Iva B\v{r}ezinov\'{a}$^2$, Sabine Andergassen$^{3,4}$, and Miriam Klopotek$^{1,5}$\\
$^1$Stuttgart Center for Simulation Science, Cluster of Excellence EXC 2075, University of Stuttgart, Universit\" atsstraße 32, 70569 Stuttgart, Germany\\
$^2$Institute for Theoretical Physics, Vienna University of Technology, Wiedner Hauptstraße 8-10, A-1040 Vienna, Austria\\
$^3$Institute of Information Systems Engineering, Vienna University of Technology, 1040 Vienna, Austria\\
$^4$Institute of Solid State Physics, Vienna University of Technology, 1040 Vienna, Austria\\
$^5$WIN-Kolleg of the Young Academy $\vert$ Heidelberg Academy of Sciences and Humanities, Karlstraße 4, 69117 Heidelberg, Germany\\
}

\maketitle

\acrodef{ode}[ODE]{ordinary differential equation}
\acrodef{2rdm}[2RDM]{two-particle reduced density matrix}
\acrodef{2hrdm}[2HRDM]{two-hole reduced density matrix}
\acrodef{3rdm}[3RDM]{three-particle reduced density matrix}
\acrodef{td2rdm}[TD2RDM]{time-dependent two-particle reduced density matrix}
\acrodef{ml}[ML]{machine learning}
\acrodef{bbgky}[BBGKY]{Bogoliubov–Born–Green–Kirkwood–Yvon}

\begin{abstract}
 Out‑of‑equilibrium quantum many‑body systems -- such as multi-electron atoms and molecules driven by strong laser fields, quenched ultracold gases, and ultrafast-excited solids -- exhibit rapid correlation buildup that underlies many emerging phenomena. Exact wave‑function methods to describe these effects scale exponentially with particle number; simpler mean‑field approaches neglect essential particle correlations. The \ac{td2rdm} formalism offers a middle ground by propagating the two‑particle density matrix and closing the \ac{bbgky} hierarchy with a reconstruction of the three‑particle cumulant, which carries information about three-particle correlations. But the validity and existence of time‑local reconstruction functionals ignoring memory effects remain unclear across different dynamical regimes.
 We show that a neural \ac{ode} model trained on exact \ac{2rdm} data (no dimensionality reduction) can reproduce its full dynamics without any explicit three‑particle information -- but only in parameter regions where the Pearson correlation between the two‑ and three‑particle cumulants is large. In contrast, in the anti‑correlated or uncorrelated regime, the neural \ac{ode} fails, indicating that no simple time‑local functional of the instantaneous two‑particle cumulant can capture the evolution. The magnitude of the time‑averaged three‑particle‑correlation buildup appears to be the primary predictor of successful extrapolation: For a moderate correlation buildup, both neural \ac{ode} predictions and existing \ac{td2rdm} reconstructions are accurate, whereas stronger values lead to systematic breakdowns. These findings pinpoint the need for memory‑dependent kernels in the three‑particle cumulant reconstruction for the latter regime.
 Our results place the neural \ac{ode} as a model‑agnostic diagnostic tool that maps the regime of applicability of cumulant expansion methods and guides the development of non‑local closure schemes. More broadly, the ability to learn high‑dimensional reduced‑density‑matrix dynamics from limited data opens a pathway to fast, data‑driven simulation of correlated quantum matter, complementing traditional numerical and analytical techniques.
\end{abstract}

\section{Introduction}

The exploration of quantum many-body systems far from equilibrium has become a central theme across ultracold-atom physics, condensed-matter theory, and ultrafast spectroscopy. Rapid experimental progress now enables the precise control of isolated atomic gases~\cite{kollath2005Spin-Charge}, the manipulation of correlated solids with intense light fields~\cite{buzzi2021Higgs-Mediated,giannetti2016Ultrafast}, the tracking of real-time electronic motion in nanostructures~\cite{driscoll2011Laser-induced} and other manipulations of quantum systems~\cite{stojchevska2014Ultrafast,caillat2005Correlated,topp2018Alloptical,vodeb2024Nonequilibrium,bedow2022Emergence,budden2021Evidence,pohl2000Towards}. These advances have revealed a broad spectrum of out-of-equilibrium phenomena (for recent reviews see e.g.~\cite{basov_towards_2017, torre_colloquium_2021, bloch_strongly_2022}), including ultrafast switches of currents in dielectrics \cite{schiffrin_optical-field-induced_2013, wachter2014Abinitio, schultze_controlling_2013, ossiander_speed_2022}, laser-induced superconductivity \cite{cremin_photoenhanced_2019, budden2021Evidence, buzzi2021Higgs-Mediated}, and ultrafast switches to hidden quantum states~\cite{stojchevska2014Ultrafast}. Parallel developments in non-equilibrium many-body theory~\cite{cazalilla2002Time-dependent,otobe2008Firstprinciples,hochstuhl2012Timedependent,hochstuhl2014Timedependent,wachter2014Abinitio,sato2018Communication,pederson2019Symplectic, eckstein2009Thermalization, schluenzen2020achieving, joost2022Dynamically, perfetto_real-time_2022, pavlyukh_time-linear_2022, lackner2015Propagating, lackner2017Highharmonic} have highlighted the need for theoretical frameworks that faithfully model (strongly) correlated dynamics of extended systems for long periods of time while remaining computationally tractable.

Accurately describing the non-equilibrium dynamics of interacting quantum many-body systems remains a central challenge because the full many-body wave function resides in a Hilbert space whose dimension grows exponentially with the particle number. Wave-function-based methods, which allow, in principle, a full parametrization of the Hilbert space such as the multiconfigurational time-dependent Hartree Fock method (MCTDHF) \cite{caillat2005Correlated, hochstuhl2014Timedependent} or matrix-product-state (MPS) techniques~\cite{Scholl2005,haegeman2016Unifying,kloss2018Timedependent,Cirac2021} provide numerically exact results but are, in general, limited to modest system sizes and short propagation times. Other approaches such as the time-dependent density-functional theory (TDDFT)~\cite{runge1984DensityFunctional,ullrich2012TimeDependent} or time-dependent Hartree–Fock (TDHF)~\cite{caillat2005Correlated} scale favorably, yet are limited to weakly (TDDFT) or uncorrelated (TDHF) systems. Several approaches exist that avoid the exponential scaling of wavefunctions by using a reduced object instead. One prominent example are methods that use non-equilibrium Green's functions (NEGF) as the central object (for an introduction see e.g.~\cite{stefanucci_2025}). However, NEGF methods typically feature a nonlinear scaling with time. Another strategy is the local-information time-evolution (LITE) approach~\cite{Artiaco2024Efficient}, which decomposes the system into subsystems at different spatial scales and evolves their reduced density matrices in parallel, systematically discarding long-range entanglement above a cutoff scale while preserving the information flow without assumptions on its microscopic details; however, its computational cost scales exponentially with the subsystem scale that controls accuracy. The \ac{td2rdm} formalism~\cite{lackner2015Propagating,lackner2017Highharmonic,donsa2023Nonequilibrium} follows a similar spirit as NEGF but features a linear scaling in time: the \ac{2rdm} $D_{12}(t)$ contains the full two-particle information, and the total energy can be expressed as a functional of $D_{12}(t)$ (if the Hamiltonian only contains one- and two-body operators), thereby reducing the computational cost from exponential to polynomial scaling. Its power stems from the explicit inclusion of dynamical two-particle correlations and from a systematic closure of the \ac{bbgky} hierarchy via a reconstruction of the \ac{3rdm} $D_{123}(t)$ as a functional of $D_{12}(t)$. The method thus relies on approximate reconstruction functionals; present state-of-the-art schemes are time-local, which guarantees the linear scaling in time of the method, and quadratic in the two-particle cumulant $\Delta_{12}(t)$, thereby neglecting memory effects and higher-order cumulants. Consequently, the accuracy of \ac{td2rdm} depends on the quality of the time-local $D_{123}$ reconstruction. We would like to point out that extensions of NEGFs methods featuring a linear scaling in time \cite{schluenzen2020achieving, joost2022Dynamically, pavlyukh_time-linear_2022} face the same problem of the necessity to reconstruct unknown components of the equations of motion in a time-local fashion.
So, the deep question is, in which regimes can we find a sufficiently accurate time-local reconstruction functional? Can numerical methods shed light on this problem?

Today, \ac{ml} introduces a diverse variety of complementary approaches to classical numerical modeling, with growing emphasis on interpretability in physics~\cite{wetzel2025interpretablemachinelearningphysics}.
Some approaches can enable integrating domain knowledge about physical laws governing the target problems of interest.
In the field of quantum physics, some examples in the existing literature model open quantum systems~\cite{Coppola2025Learning,zhang2025Neural,liu2023Predicting,Mazza2021,Chen2022,Chen2025}, investigate quantum-information tasks~\cite{Chen2022,Genois2021}, measure entanglement~\cite{huang2025Direct}, predict the phonon blockade effect~\cite{Zeng2021Application}, use neural quantum states to overcome the exponential scaling in the Hilbert space~\cite{Lange2024architectures,vandewalle2024ManyBody}, combine \ac{ml} with quantum computing~\cite{ye2025Simulating}, and enable quantum (optimal) control~\cite{Chen2025Optimal,Norambuena2024,Brown2021Reinforcement,Zeng2020Quantum}.

Dynamical processes are generally governed by differential equations and have been subject to numerous investigations using \ac{ml}~\cite{Kidger2020a,Toth2020Hamiltonian,MorrillSKF21,Dietrich2022,Carnazza2022,Carnazza2024,Schmitt2020,Lin2022,Zhao2024,Mohseni2024,Zeng2024,An2025,Kaneko2025,Sun2025,Schmitt2025,Cemin2025}.
In the present study, we focus specifically on systems whose time evolution can be described by \acp{ode}.
Neural \acp{ode} offer a powerful venue of thinking because they combine a rudimentary simulation of propagation of time through a differential equation, with data-driven ML~\cite{chen2018Neural}. It is a promising candidate for modeling diverse time-dependent datasets in and out of science~\cite{qian2021Integrating,worsham2025guide,hoege2022Improving,hananeh2021Beyond,lee2021Parameterized,niu2024applications,sorourifar2023PhysicsEnhanced,bram2024Lowdimensional}. To our knowledge, its application in classical or quantum many-body physics is scarce. Neural \acp{ode} have been applied to reduced representations of quantum dynamics~\cite{choi2022Learning}, holographic quantum chromodynamics~\cite{Hashimoto2021}, or to the Hamiltonian learning problem~\cite{Heightman2025Solving}. Other \ac{ml} approaches have been used to investigate Markovian and non-Markovian dynamics or noise of quantum systems~\cite{Nelson2022,Mukherjee2024Noise}.

Testing for the Markov property in time series is a longstanding problem in many fields such as statistics, molecular dynamics, and quantum information theory. Classical statistical approaches based on kernel smoothing~\cite{aitsahalia1999Interest} or local polynomial regression of the conditional characteristic function~\cite{Chen2012Testing} suffer from the curse of dimensionality and become unreliable in moderate- to high-dimensional settings. More recent proposals employ deep conditional generative models such as mixture density networks~\cite{zhou2023Testing} or random-forest-based doubly robust procedures~\cite{shi2020Does}, yet all of these methods require the time series to be strictly stationary, i.e.~the joint probability distribution of the process is time-invariant, and assume a stochastic data-generating process. A complementary, model-free approach for single-molecule trajectories has been proposed by Berezhkovskii and Makarov~\cite{Berezhkovskii2018SingleMolecule}: by comparing transition-path probabilities $P(a \to b|x)$ with the value $1/4$ expected for one-dimensional Markov diffusion, it detects non-Markovian behavior without fitting the data to any model. However, this criterion is restricted to one-dimensional reaction coordinates and relies on equilibrium sampling. In the quantum-information context, Wolf et al.~\cite{wolf2008Assessing} introduced computable criteria and a continuous measure of Markovianity for quantum channels, deciding whether a given completely positive trace-preserving map is consistent with a Lindblad master equation. While this provides a rigorous framework for open quantum systems, it requires full quantum process tomography and operates within the Lindblad formalism; it is therefore not applicable to the reduced dynamics of a closed system. The trace-distance-based information-backflow criterion~\cite{Bylicka2017Constructive} was recently applied by Banerjee~\cite{Banerjee2025NonMarkovianity} to subsystem dynamics in a quenched Ising chain, detecting non-Markovianity of small subsystem reduced density matrices. However, this approach also relies on the open-system framework and scales exponentially with subsystem size, making it impractical beyond a few lattice sites. In contrast, neural \acp{ode} impose no stationarity, probabilistic, or open-system assumptions. Moreover, because they learn a time-local vector field by construction, their predictive accuracy provides a continuous, physically interpretable measure of the degree of Markovianity, directly informing the development of closure functionals for reduced-order many-body methods.

In this paper, we embark on a study of quantum many-body systems out of equilibrium, viewed through \acp{2rdm}~\cite{donsa2023Nonequilibrium}. For this purpose, we investigate the dynamics of the paradigmatic Fermi-Hubbard model \cite{Qin2021}.
Depending on the parameters, it shows a broad range of dynamical regimes, reaching from weakly to strongly correlated, and from weak to strong excitation by tuning only two parameters.
In the present work, we intentionally focus on small system sizes and comparatively simple lattice settings, since exact-reference data generation becomes rapidly more expensive with increasing system size and dimensionality.
Within the computationally feasible regime, however, the study is already broad: we analyze over 300 distinct parameter configurations, each showing different dynamics and dynamical regimes.
We demonstrate that neural \acp{ode} offer interesting insights: We find indications that time-local reconstructions of the \ac{3rdm} may only be possible in certain regimes in the parameter space. The infusion of physics knowledge is neither straightforward nor does prior physical information necessarily lead to better results.

The structure of the paper is as follows: In Sec.~\ref{sec:methods} we briefly introduce the quenched Fermi-Hubbard model, which will be the system under investigation. We describe the equations of motion for the \ac{2rdm} and concisely revisit the cumulant reconstruction needed to close them. The neural \ac{ode} model used in this investigation is presented together with the data preprocessing. In Sec.~\ref{sec:results}, we first identify some structure in the parameter space of the Fermi-Hubbard model. This is used to investigate the performance of the neural \ac{ode} across the parameter plane. To improve the stability of the prediction, we enforce physical constraints in the loss. Last, we discuss the results and show perspectives for future investigations in Sec.~\ref{sec:discussion}.

\section{Methods}
\label{sec:methods}

\subsection{Non-equilibrium Fermi-Hubbard model}

\begin{figure}[t]
    \centering
    \includegraphics[width=\textwidth]{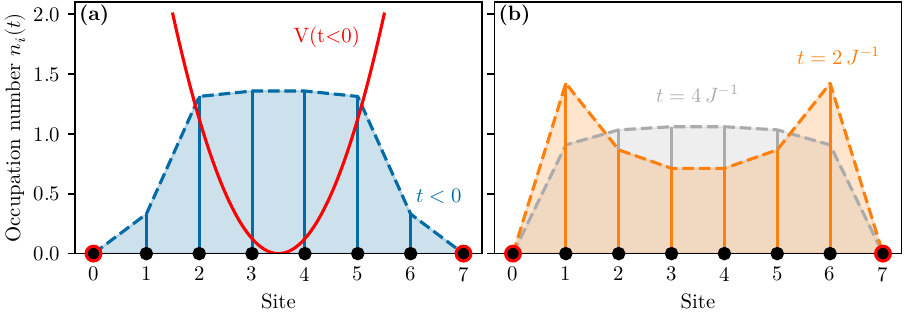}
    \caption{Fermi-Hubbard model consisting of $6$ sites, initially ($t < 0$) subject to an external harmonic trap. Dirichlet (hard-wall) boundary conditions on site $0$ and $7$ are indicated by red circles. (a) The one-particle site-occupation numbers $n_i$ of the trapped ground state, which correspond to an excited configuration of the unconfined system after the quench. (b) Temporal evolution of the occupations $n_i (t)$ after the quench, displayed at $t = 2\,J^{-1}$ and $t = 4\,J^{-1}$. The parameters of Eq.~\eqref{eq:fermi-hubbard} are $V = J$ and $U = J$.}
    \label{fig:fermi-hubbard}
\end{figure}

We consider a one‑dimensional Fermi–Hubbard chain comprising $M_s$ physical sites together with two auxiliary sites labeled $0$ and $M_s + 1$ that are fixed to be unoccupied throughout the dynamics. Dirichlet boundary conditions are imposed at these auxiliary sites, a choice that mirrors the confinement used in quantum‑simulator experiments with trapped ultracold atoms (see, e.g., Ref.~\cite{bloch2012Quantum}). While Dirichlet boundaries are natural for the experimental setting, alternative boundary conditions (e.g., periodic) can be incorporated straightforwardly within the same formalism. The initial state is prepared in a harmonic trapping potential, which is abruptly switched off at $t=0$, after which the system evolves under the standard Fermi-Hubbard Hamiltonian~\cite{donsa2023Nonequilibrium}. The full Hamiltonian in second quantization is given by
\begin{align}
    \mathcal{H} = - J \sum_{\langle i,j \rangle} \sum_\sigma a^\dagger_{i \sigma} a_{j \sigma} + U \sum_i n^\uparrow_i n^\downarrow_i + \sum_{i,\sigma} V_i (t) a^\dagger_{i \sigma} a_{i \sigma} \,.
    \label{eq:fermi-hubbard}
\end{align}
Here, $J$ denotes the hopping amplitude, $\langle i,j\rangle$ indicates nearest-neighbor pairs, $a_{i\sigma}^{(\dagger)}$ are the single-particle annihilation (creation) operators, $n_i^{\uparrow(\downarrow)} = a_{i\uparrow(\downarrow)}^\dagger a_{i\uparrow(\downarrow)}$ are the corresponding spin-resolved occupation number operators at site $i$, and $U$ characterizes the on-site interaction strength governing the interaction energy. The system before and after the quench can be seen in Fig.~\ref{fig:fermi-hubbard}.

For the initial trap, we use a harmonic potential
\begin{align}
    V_i (t) = \theta (-t) \frac{V^2}{2} \left(i - \frac{M_s + 1}{2} \right)^2\,,
\end{align}
which initially traps the system in the ground state of the harmonic trap and is turned off at $t=0$. This potential quench, parametrized by $V$, induces the dynamics. Although this is a quite simple model, it is still complex enough to show rich dynamics.

In the following analysis, we study the spin‑symmetric Fermi–Hubbard model at half filling, i.e., with a total particle number $N = M_s$ and equal numbers of up‑ and down‑spin particles $N_\uparrow=N_\downarrow$, corresponding to a total spin‑singlet configuration.

\subsection{Time-evolution of the 2RDM}

A $p$-particle reduced density matrix (pRDM) is obtained from the $N$‑body state $|\Psi(t)\rangle$ by tracing over all but $p$ particles
\begin{equation}\label{eq:pRDM}
    D_{1\ldots p}(t) = \binom{N}{p}p! \text{Tr}_{p+1\ldots N} |\Psi (t)\rangle\langle \Psi (t)|,
\end{equation}
with $\binom{N}{p}p!$ being the chosen normalization for the pRDMs. This particular normalization has the advantage that the cumulant decompositions (see Eq.~\ref{eq:2RDM-cumulant} and Eq.~\ref{eq:3RDM-cumulant} below) do not have any prefactors. Correspondingly, the \ac{2rdm} is given by
\begin{equation}
    D_{12}(t) = N(N - 1) \text{Tr}_{3\ldots N} |\Psi (t)\rangle\langle \Psi (t)|\,.
\end{equation}
The second equation in the \ac{bbgky} hierarchy, i.e., the equation of motion for the 2RDM,
\begin{equation}
    i \partial_t D_{12} = [h_1 + h_2 + W_{12}, D_{12}] + \text{Tr}_3[W_{13} + W_{23}, D_{123}]\,,
  \label{eq:eom-2RDM}
\end{equation}
contains the single‑particle Hamiltonian $h_i$, the two‑body interaction $W_{ij}$, and the \ac{3rdm} $D_{123}$. $\text{Tr}_3$ denotes the trace over the third particle. Closing Eq.~\eqref{eq:eom-2RDM} therefore requires a reconstruction functional $D^{\text{R}}_{123}[D_{12}]$. This leads to the \ac{td2rdm} theory introduced in Refs.~\cite{lackner2015Propagating, lackner2017Highharmonic, donsa2023Nonequilibrium}.

In a total‑spin‑singlet configuration, as considered here, the evaluation of Eq.~\eqref{eq:eom-2RDM} requires only the propagation of the mixed‑spin block $\langle j_1\uparrow j_2\downarrow| D_{12}(t)|i_1\uparrow i_2\downarrow\rangle = D^{j_1\uparrow j_2\downarrow}_{i_1\uparrow i_2\downarrow}(t)$, with $|i_1\uparrow i_2\downarrow\rangle$ two-particle product states of spin orbitals, since it already encodes the complete information of the full \ac{2rdm} $D_{12}(t)$ \cite{lackner2015Propagating}. For completeness, we note that elements of the 2RDM can also be obtained in the second quantization formalism through $D^{j_1\uparrow j_2\downarrow}_{i_1\uparrow i_2\downarrow}(t) = \langle \Psi(t)|a_{i_1\uparrow}^\dagger a_{i_2\downarrow}^\dagger a_{j_2\downarrow} a_{j_1\uparrow}| \Psi(t)\rangle$. All other spin components can be recovered from this block by straightforward exchange or spin‑flip symmetry relations.

\subsubsection{Cumulant expansion}
\label{sec:cumulant-expansion}
Any $p$RDM can be decomposed into antisymmetrised products of lower‑order RDMs plus a connected part (the cumulant)~\cite{kutzelnigg1999Cumulant}. For the \ac{2rdm} the decomposition reads
\begin{equation}
  D_{12}= \hat A D_{1} D_{2} +\Delta_{12} ,
  \label{eq:2RDM-cumulant}
\end{equation}
where $\hat A$ antisymmetrises the product and $\Delta^{(2)}_{12}$ is the two‑particle cumulant. For the \ac{3rdm} one obtains
\begin{equation}
 D_{123}= \hat{A} D_{1} D_{2} D_{3} +\hat{A} \Delta_{12} D_{3} +\Delta_{123} \,,
\label{eq:3RDM-cumulant}
\end{equation}
where $\Delta_{123}$ is the three‑particle cumulant (the genuinely connected three‑body correlation). We have chosen a convention for $\hat A$ such that every topologically distinct term contributes only ones in order to avoid normalization in the cumulant expansion Eq.~\eqref{eq:3RDM-cumulant}. The first term contains only products of the 1RDM so that the three particles are uncorrelated in their dynamics.
The second term shows that once the two‑particle cumulant $\Delta_{12}$ is known, all pairwise correlations in the three‑body system are already determined; only the last term $\Delta_{123}$ adds the correlated three‑body part. As has been shown in Ref.~\cite{lackner2015Propagating}, parts of the three-particle cumulant $\Delta_{123}$ can be exactly reconstructed from the \ac{2rdm} through the knowledge of its traces. The irreducible three-body part of $\Delta_{123}$ denoted as $\Delta_{123,\text{K}}$ with $\text{Tr}_3\Delta_{123,\text{K}}=0$, however, cannot be expressed through lower‑order RDMs.
Consequently, an accurate reconstruction of $D_{123}$ depends on a good approximation of $\Delta_{123,\text{K}}$, while the bulk of the many‑body information is already encoded in $D_{12}$ (through $\Delta_{12}$).

All practical \ac{td2rdm} implementations close the \ac{bbgky} hierarchy by replacing the \ac{3rdm} $D_{123}$ with a functional of the \ac{2rdm} $D_{12}$ (i.e.\ of the two-particle cumulant $\Delta_{12}$). All presently employed approximations are time‑local, thereby neglecting any memory dependence. A more detailed derivation and specific functional forms can be found in Ref.~\cite{donsa2023Nonequilibrium}.

\subsection{Neural \ac{ode}}

Neural \acp{ode} provide a data‑driven framework to learn a continuous‑time dynamical system from sampled trajectories~\cite{chen2018Neural}. Given a set of discrete \ac{2rdm} snapshots $\{D^{j_1\uparrow j_2\downarrow}_{i_1\uparrow i_2\downarrow}(t_k)\}_{k=0}^{K}$ obtained from the propagation of the wavefunction through exact diagonalization of the Schrödinger equation, a neural \ac{ode} seeks a parametric vector field $\mathcal{F}_\theta\bigl(D^{j_1\uparrow j_2\downarrow}_{i_1\uparrow i_2\downarrow}\bigr)$ such that
\begin{equation}
    \frac{\mathrm{d}}{\mathrm{d}t}D^{j_1\uparrow j_2\downarrow}_{i_1\uparrow i_2\downarrow}(t)=\mathcal{F}_\theta\bigl(D^{j_1\uparrow j_2\downarrow}_{i_1\uparrow i_2\downarrow}(t)\bigr)\,,
\end{equation}
where $\theta$ denotes the trainable network parameters. The loss function is the integrated discrepancy between the network prediction and the reference data,
\begin{equation}
    \mathcal{L}(\theta)=\frac{1}{K}\sum_{k=1}^{K}\bigl\| \Phi_\theta(t_k;D^{j_1\uparrow j_2\downarrow}_{i_1\uparrow i_2\downarrow}(t_0)) - D^{j_1\uparrow j_2\downarrow}_{i_1\uparrow i_2\downarrow}(t_k)\bigr\|_F^{2},
    \label{eq:neural-ode-loss}
\end{equation}
with $\Phi_\theta$ the solution of the neural ODE obtained by an adaptive ODE solver and $\|\cdot\|_F$ the Frobenius (Schatten‑2) norm.
After training, the learned vector field can be evaluated on unseen initial conditions to test whether the dynamics is well described by a time‑local functional of the instantaneous \ac{2rdm}.

Thus, neural \acp{ode} serve as an independent, model‑agnostic diagnostic tool to map the parameter regimes (in the $U$–$V$ plane) where the \ac{td2rdm} closure can be reliably approximated by time‑local reconstructions as discussed in Sec.~\ref{sec:cumulant-expansion}.

In practice, we employ a fully connected architecture with two hidden layers, each with a dimension of $2048$. We tested deeper and wider feed-forward neural \ac{ode} variants, but did not observe meaningful differences in predictive behavior, hence we stick to this architecture. The mixed-spin block of the \ac{2rdm} $D^{j_1\uparrow j_2\downarrow}_{i_1\uparrow i_2\downarrow}$ contains $1296$ complex-valued entries. To pass them through the neural \ac{ode}, we must separate the real and imaginary parts, yielding a total of $2592$ real values. This can be reduced by leveraging its hermiticity and taking only the upper triangular values, which in turn results in $1296$ real values. For each parameter configuration in the $U$-$V$ plane, we divide our time series into training, validation, and test sets. The first $3000$ time steps are used for training and the next $1000$ time steps are used for validation, where one time step is $\Delta t = 0.01 \, J^{-1}$. The remainder can be used for testing. We apply min-max normalization to the real and imaginary parts separately to scale them to $[0,1]$. Here, we first normalize the train set, then use the same normalization constants for both the validation and test sets. We use the Python library \texttt{torchdiffeq} \cite{torchdiffeq} for the implementation of the neural \ac{ode}.

It is important to mention that, compared to other applications of neural \acp{ode}, which typically reduce high-dimensional data to a lower-dimensional manifold and then train the neural \ac{ode} on the low-dimensional representation~\cite{guo2025Efficient,lai2022Neural,norcliffe2021Neural,sholokhov2023Physicsinformed,lee2021Parameterized}, we train the model directly on the high-dimensional data. We intentionally do not reduce the dimensionality of the data, since we want to test the Markovianity of the dynamics. If we were to reduce the dimensionality, and then train the neural \ac{ode} on the low-dimensional representation, we would not be able to distinguish whether a failure of the neural \ac{ode} is due to a non-Markovian dynamics or due to information loss in the dimensionality reduction, resulting in a non-Markovian dynamics in the low-dimensional representation. By training the neural \ac{ode} directly on the high-dimensional data, we can directly test the Markovianity of the dynamics without this ambiguity. Furthermore, the size of the training set is of the same order of magnitude as the dimensionality. To our best knowledge, this is the first time that neural \acp{ode} have been trained in this way, which, if successful, could open up new application possibilities.

\section{Results}
\label{sec:results}

\subsection{Time-local reconstruction for different parameter configurations}

To determine where a time‑local reconstruction of the three‑particle cumulant $\Delta_{123}(t)$ is justified, we first analyze three quantitative indicators obtained from the exact dynamics of the two‑particle cumulant, $\Delta_{12}(t)$, and the kernel component of the three‑particle cumulants, $\Delta_{123, \text{K}}(t)$, for a broad range of interaction strengths $U$ and quench amplitudes $V$. These measures serve both to interpret the neural \ac{ode} results and to determine the general structure of the $U$-$V$ parameter plane.

\begin{figure}[t]
    \centering
    \includegraphics[width=\textwidth]{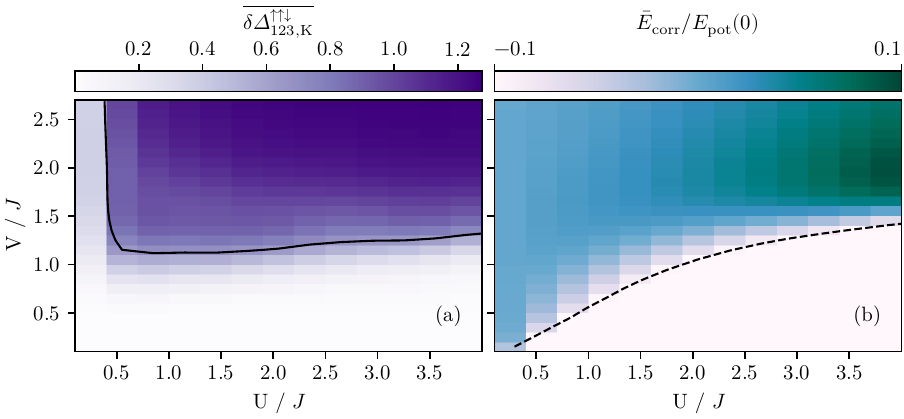}
    \caption{We study the six-site Fermi-Hubbard model of Eq.~\eqref{eq:fermi-hubbard}.
    Both plots show a parameter scan over the initial potential strength $V$ and the interaction $U$. The dynamical buildup of three-particle correlations as described in Eq.~\eqref{eq:correlation-buildup} is visualized in (a). The solid black line corresponds to $\overline{\delta\Delta^{\uparrow\uparrow\downarrow}_{123,\text{K}}} = 0.65$, which indicates the crossover between moderate and strong correlation buildup. In (b) the time-averaged correlation energy relative to the initial excitation energy $\bar{E}_{\text{corr}}/E_{\text{pot}}(0)$ is shown, where the black dashed line corresponds to $\bar{E}_{\text{corr}}/E_{\text{pot}}(0) = -0.1$.
    This delimits the regime of strong two-particle correlation buildup. The color-bar is cut below $\bar{E}_{\text{corr}}/E_{\text{pot}}(0) = -0.1$ to have a clear distinction between the two regimes.
    Both values were chosen in accordance with Fig.~\ref{fig:correlations_performance} since they also delimit the positive and negative as well as the strong and weak correlations of the cumulant dynamics, respectively.}
    \label{fig:cumu-buildup_energy}
\end{figure}

\begin{figure}[t]
    \centering
    \includegraphics[width=\textwidth]{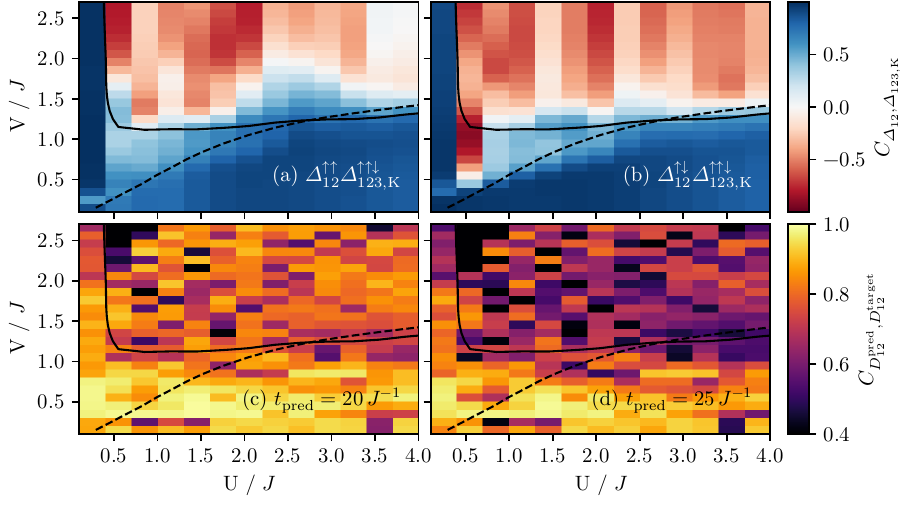}
    \caption{In (a) and (b) the Pearson correlation coefficient $C_{\Delta_{12},\Delta_{123,\text{K}}}$ between the Frobenius norms of the kernel component of the three-particle cumulant $\Delta_{123,\text{K}}^{\uparrow\uparrow\downarrow}$ and the two-particle cumulants $\Delta_{12}^{\uparrow\uparrow}$ and $\Delta_{12}^{\uparrow\downarrow}$, respectively, are shown across the $U$-$V$ plane for the six-site Fermi-Hubbard model of Eq. \eqref{eq:fermi-hubbard}. The contour lines $\overline{\delta\Delta^{\uparrow\uparrow\downarrow}_{123,\text{K}}} = 0.65$ and $\bar{E}_{\text{corr}}/E_{\text{pot}}(0) = -0.1$ from Fig.~\ref{fig:cumu-buildup_energy} are visualized as well.
    In (c) and (d) the Pearson correlation coefficient between the predicted and the exact time series of the \ac{2rdm} $C_{D_{12}^\text{pred},D_{12}^\text{target}}$ is visualized for a prediction length of $t_\text{pred} = 20 \, J^{-1}$ and $t_\text{pred} = 25 \, J^{-1}$ (2000 and 2500 time steps), respectively, starting from $D_{12} (t=40\,J^{-1})$. For each parameter configuration, one model was trained.}
    \label{fig:correlations_performance}
\end{figure}

The net increase of genuine three‑particle correlations in the irreducible component during the evolution is quantified by
\begin{equation}
    \overline{\delta\Delta^{\uparrow\uparrow\downarrow}_{123,\text{K}}} = \frac{1}{T} \int_0^T \left\| \Delta^{\uparrow\uparrow\downarrow}_{123,\text{K}} (t) \right\|_F - \left\| \Delta^{\uparrow\uparrow\downarrow}_{123,\text{K}} (0) \right\|_F \text{d}t \,,
    \label{eq:correlation-buildup}
\end{equation}
where $\|\cdot\|_F$ denotes the Frobenius  norm. The contour $\delta\Delta_{123,\text{K}}=0.65$, which can be used to separate regimes of moderate ($\delta\Delta_{123,\text{K}}\le0.65$) and strong ($\delta\Delta_{123,\text{K}}>0.65$) three‑particle correlations buildup, as visible in Fig.~\ref{fig:cumu-buildup_energy}(a), can be simultaneously used to delimit regimes of positive and negative correlations in the dynamics of the cumulants, see Fig.~\ref{fig:correlations_performance}. For the upper time limit we take $T = 50\, J^{-1}$ for the whole parameter scan.

As another measure of correlations buildup, we use the instantaneous two‑particle correlation energy defined as
\begin{equation}
    E_\text{corr} (t) = \operatorname{Tr}_{12} W_{12} \Delta_{12} (t) \,,
\end{equation}
which in the case of the Fermi-Hubbard model reduces to
\begin{equation}
    E_\text{corr} (t) = U \sum_j \Delta^{j\uparrow j\downarrow}_{j\uparrow j\downarrow} (t) \,.
\end{equation}
We compare the time–averaged correlation energy
\begin{equation}
    \bar{E}_{\text{corr}}=\frac{1}{T}\int_{0}^{T} E_{\text{corr}}(t)\,\text{d}t\,,
\end{equation}
with the initial potential energy of the quenched state
\begin{equation}
    E_\text{pot}(0) = \operatorname{Tr} D_1 (t=0) V_1 (t=0) \,
\end{equation}
in Fig.~\ref{fig:cumu-buildup_energy}(b). The color scale is truncated at $-0.1$ for visual clarity; values below this threshold are present but not displayed.  Throughout the explored parameter space, the correlation energy is initially negative and increases during the time evolution, attaining positive average values for some parameter configurations. Figure~\ref{fig:cumu-buildup_energy}(b) thus visualizes the buildup of two‑particle correlations. The cut‑off at $\bar E_\text{corr}/E_\text{pot}(0)=-0.1$ was again chosen empirically to separate regimes of pronounced correlation‑energy increase from those exhibiting only a modest rise.
Both contour lines in Fig.~\ref{fig:cumu-buildup_energy} indicate structures in the dynamics of the \ac{2rdm} across the parameter plane, which later help to delimit regimes of positive and negative correlations in the dynamics of the cumulants.

We hypothesize that in regions with high (anti-)correlation between two-particle and three-particle cumulants, the neural \ac{ode} should perform well: The information about the three-particle cumulant, which is required to close the equations of motion for the \ac{2rdm}, should be contained in the time-local \ac{2rdm}. In the correlated regime, this assumption seems natural. In the anti-correlated regime, there is a strong correlation between these two quantities, though negative.
This could hint at the existence of a time-local reconstruction functional, which takes this inverse relation into account. The current reconstruction functionals only perform well in the positively correlated regime, since they are proportional to polynomials of the two-particle cumulant. The existence of such an inverse functional would allow for efficient numerical integration of the equations of motion with accurate results even in the regime of large quenches and large correlation buildups. The neural \ac{ode} serves as a model-agnostic tool: It can only learn Markovian dynamics by design. Thus, if memory is required to reconstruct the three-particle cumulant and hence close the equations of motion, the model should not be able to learn and predict the exact dynamics of the \ac{2rdm}.

To assess this hypothesis, we look at the Pearson correlation coefficient $C_{f,g}$~\cite{pearson1896Mathematical}, which provides a quantitative measure of the linear relationship between two time‑dependent observables $f(t)$ and $g(t)$. It is defined as the normalized covariance
\begin{equation}
C_{f,g}= \frac{\displaystyle\frac{1}{T-t_0}\int_{t_0}^{T}\left[f(t)-\bar f\right]\left[g(t)-\bar{g}\right]\text{d}t}
{\sigma_f\sigma_g}\,,
\label{eq:pearson}
\end{equation}
where $\bar f$ and $\bar g$ denote the temporal averages
\begin{equation}
    \bar{f}=\frac{1}{T-t_0}\int_{t_0}^{T} f(t)\text{d}t\,,\qquad
    \bar{g}=\frac{1}{T-t_0}\int_{t_0}^{T} g(t)\text{d}t\,,
\end{equation}
and $\sigma_f$, $\sigma_g$ are the corresponding standard deviations
\begin{equation}
    \sigma_f=\sqrt{\frac{1}{T-t_0}\int_{t_0}^{T}\left[f(t)-\bar f\right]^2{\rm d}t}\,,
    \qquad
    \sigma_g=\sqrt{\frac{1}{T-t_0}\int_{t_0}^{T}\left[g(t)-\bar g\right]^2{\rm d}t}\,.
\end{equation}
By construction, $-1\le C_{f,g}\le1$; $C_{f,g}=1$ ($-1$) indicates perfect (anti‑)correlation, while $C_{f,g}=0$ corresponds to the absence of correlation. In the present work, $C_{f,g}$ is employed to assess the temporal correlation between the size of the two‑particle cumulant $\Delta_{12}(t)$ and the size of the kernel component of the three‑particle cumulant $\Delta_{123,\text{K}}(t)$ as measured by the Frobenius norm. Since the initial rise of the cumulants is always correlated and we are only interested in the correlations of the steady-state fluctuations, we omit the start of the time series and empirically choose $t_0 = 10\,J^{-1}$. According to our hypothesis, a high absolute value of $C_{\Delta_{12},\Delta_{123,\text{K}}}$ supports the validity of time‑local reconstruction functionals, whereas low absolute values signal the need for memory‑dependent extensions.
In Fig.~\ref{fig:correlations_performance}(a) and (b) the Pearson correlation coefficient between the Frobenius norms of the kernel component of the three-particle cumulant $\Delta_{123,\text{K}}^{\uparrow\uparrow\downarrow}$ and the two-particle cumulants $\Delta_{12}^{\uparrow\uparrow}$ and $\Delta_{12}^{\uparrow\downarrow}$, respectively, are shown. Both contour lines, which are equal to those in Fig.~\ref{fig:cumu-buildup_energy}, help to delimit the crossover between positive and negative correlation between the cumulant dynamics. Note that evaluating in Fig.~\ref{fig:cumu-buildup_energy} and \ref{fig:correlations_performance} the entire three-particle cumulant instead of only the kernel component (such as in Ref.~\cite{donsa2023Nonequilibrium}) leads to almost the same results because the trace information is a small, practically time-independent contribution.

We use $D_{12}(t_0 = 40\,J^{-1})$ as the initial input to the neural \ac{ode}, which is then employed to predict the subsequent evolution of the \ac{2rdm} starting from $t_0 = 40\,J^{-1}$. Consequently, we define the prediction length as
\begin{equation}
    t_\text{pred} = t - t_0\,.
    \label{eq:prediction-length}
\end{equation}
To quantify the predictive performance of the neural \ac{ode}, we calculate the Pearson correlation coefficient between the predicted time series of the \ac{2rdm} and the actual time series for the reduced representation (upper triangular values only) $C_{D_{12}^\text{pred},D_{12}^\text{target}}$ for a prediction length of $t_\text{pred} = 20 \, J^{-1}$ and $t_\text{pred} = 25 \, J^{-1}$ (2000 and 2500 time steps) starting from $D_{12}(t=40\,J^{-1})$, visualized in Fig.~\ref{fig:correlations_performance}(c) and (d), respectively. The contour lines are the same as in Fig.~\ref{fig:cumu-buildup_energy}. For a prediction length of $t_\text{pred} = 20\,J^{-1}$ the performance is high in the correlated regime and reasonably good in the anti-correlated regime. However, for a longer prediction length of $t_\text{pred} = 25 \, J^{-1}$, the performance in the anti-correlated regime deteriorates significantly, while it remains good in the correlated regime. Notably, in the anti-correlated region, values of the Pearson coefficient $C_{D_{12}^\text{pred},D_{12}^\text{target}} \in [0.4,0.7]$ are observed. This means that time-local neural \ac{ode} still captures a substantial part of the dynamics in these regions, which suggests that the non-Markovian memory effects are weak or short-lived. In such parameter regions, reconstruction functionals that incorporate only a short memory window may already suffice for an accurate closure of the \ac{bbgky} hierarchy.

The Pearson correlation coefficient is used only as a post-processing metric to evaluate how accurately the neural \ac{ode} reproduces the \ac{2rdm} time series. The model is trained by minimizing the mean squared error between predicted and reference trajectories, as defined in Eq.~\eqref{eq:neural-ode-loss}. This training loss decreases by several orders of magnitude for the entire parameter scan, such that we regard the optimization as successful for all $U-V$ values. Given the successful training, we then use the Pearson coefficient as a diagnostic for effective Markovianity: since neural \acp{ode} learn time-local (Markovian) dynamics, the coefficient should be high when the underlying dynamics is effectively Markovian, and lower when non-Markovian memory effects are relevant.

To further investigate the accuracy of the prediction, we calculate two quantities from the predicted time series, namely the occupation number $n (t)$ and the doublon occupation number $d (t)$ (number of pairs on a site).
\begin{figure}[t]
    \centering
    \includegraphics[width=\textwidth]{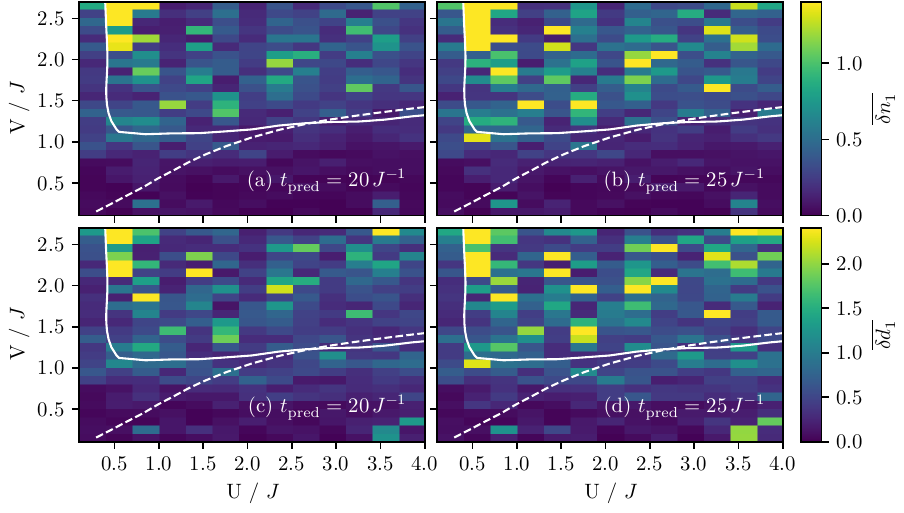}
    \caption{The upper plots show the time-integrated deviations of the occupation number of site 1 as described in Eq. \eqref{eq:occupation_relative_error} between the predicted and the exact occupation number over time for a prediction length of (a) $t_\text{pred} = 20 \, J^{-1}$ and (b) $t_\text{pred} = 25 \, J^{-1}$. Similarly, in (c) and (d) the same quantity is illustrated for the doublon occupation number $d_1$ of site 1. The contour lines $\overline{\delta\Delta^{\uparrow\uparrow\downarrow}_{123,\text{K}}} = 0.65$ and $\bar{E}_{\text{corr}}/E_{\text{pot}}(0) = -0.1$ from Fig.~\ref{fig:cumu-buildup_energy} are visualized as well. For each parameter configuration, one model was trained, with the prediction starting from $D_{12} (t=40\,J^{-1})$.}
    \label{fig:deviation}
\end{figure}
Since all sites show similar dynamics, we only consider site 1. The accuracy of the prediction is determined by the time-integrated deviations from the exact result
\begin{align}
\overline{\delta n_1} = \frac{\displaystyle\frac{1}{T}\int_{t_0}^{t_0 + T}\left|n_{1}^{\text{target}}(t) - n_{1}^{\text{predicted}}(t)\right| \text{d}t}{\displaystyle\frac{1}{T}\int_{t_0}^{t_0 + T}n_{1}^{\text{target}}(t) \text{d}t} \,.
\label{eq:occupation_relative_error}
\end{align}
This quantity as well as the deviations for the doublon occupation number (i.e.~occupation of sites by two-particles) $\overline{\delta d_1}$ are shown in Fig.~\ref{fig:deviation} for both prediction lengths. Again, we can see that the error is low in the correlated regime, while it is overall higher (and more strongly fluctuating) in the anti-correlated regime. Figure~\ref{fig:deviation} also depicts that even for a shorter prediction length, there is a significant difference in the accuracy of the prediction in the correlated and anti-correlated regimes.

These results contradict our hypothesis and thus suggest that an accurate time-local reconstruction functional cannot be found in the anti-correlated regime. The solid contour line separates moderate from strong correlations buildup of the kernel component of the three-particle cumulant ($\overline{\delta\Delta^{\uparrow\uparrow\downarrow}_{123,\text{K}}} = 0.65$). In both Fig.~\ref{fig:correlations_performance} and \ref{fig:deviation}, this line quite accurately localizes regions of good and bad performance, indicating that the existence of a time-local reconstruction functional depends on this quantity.

Training one model for one parameter configuration in the $U$-$V$ plane took approximately 6 hours on an NVIDIA A100 GPU. Due to the high computational cost and the huge amount of parameter configurations, we were able to train only one model per configuration. This explains the fluctuations in the performance across the $U$-$V$ plane in Fig.~\ref{fig:correlations_performance} and \ref{fig:deviation}, which could be reduced by either training an ensemble of models or performing hyper-parameter optimization for each parameter configuration. Even though these fluctuations exist, the trends in performance in the different regimes described above are clearly discernible.
\begin{figure}[t]
    \centering
    \includegraphics[width=\textwidth]{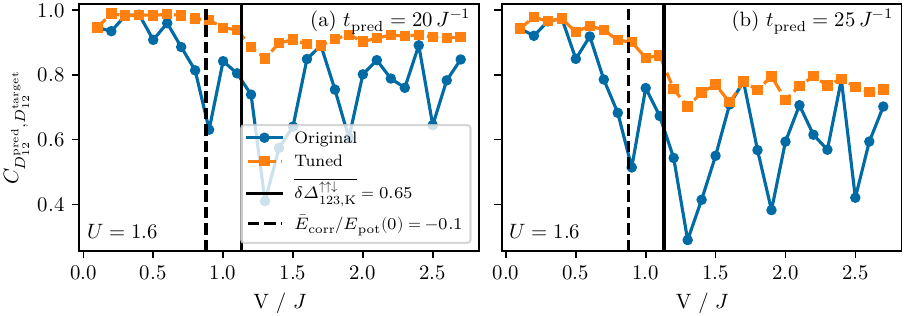}
    \caption{Line scan at fixed $U=1.6 \, J$, with the Pearson correlation coefficient $C_{D_{12}^{\text{pred}}, D_{12}^{\text{target}}}$ as a function of $V$. Comparison between the original prediction from Fig.~\ref{fig:correlations_performance} and the prediction obtained by a hyper-parameter-optimized model at each $V$. The hyper-parameter-optimization clearly suppresses fluctuations while preserving the overall trend. The dashed and solid vertical lines correspond to the intersections with the contour lines $\bar{E}_{\text{corr}}/E_{\text{pot}}(0) = -0.1$ and $\overline{\delta\Delta^{\uparrow\uparrow\downarrow}_{123,\text{K}}} = 0.65$, respectively, from Fig.~\ref{fig:cumu-buildup_energy}.}
    \label{fig:line-scan-tuned}
\end{figure}
To investigate whether hyper-parameter optimization reduces the fluctuations, we fixed $U=1.6 \, J$ and performed a line scan over $V$. For each value of $V$, we performed hyper-parameter optimization. Figure~\ref{fig:line-scan-tuned} compares the original result from Fig.~\ref{fig:correlations_performance} with the optimized result. The optimization substantially reduces fluctuations and supports the robustness of the trends discussed above.
Since the Pearson correlation coefficient slightly decreases towards the dashed line, similar to the correlation between the cumulants in Fig.~\ref{fig:correlations_performance}(b), this suggests that the dynamics is Markovian to a good approximation in the correlated regime and memory is required as soon as this correlation decreases. For longer prediction horizons even small deviations from Markovianity can accumulate, which explains the shift between Fig.~\ref{fig:line-scan-tuned} (a) and (b). It is remarkable that the hyper-parameter optimization eliminates the noise in the prediction, making the tuned model less uncertain.

The training of the neural ODE based on the full data as performed in this study is computationally costly. While it is not within the scope of the present study to focus on the computational cost, there are some obvious strategies for cost reduction: (i) reducing the dimensionality of the input data and (ii) reducing the model size. However, we intentionally do not prioritize these strategies in the present work: dimensionality reduction may discard information that is crucial for assessing Markovianity, and therefore could bias the main diagnostic objective. Likewise, substantially smaller networks introduce a stronger information bottleneck, which is not desired in our setup because it can limit representational capacity and confound the interpretation of memory effects.

\subsection{Stability of the prediction}

\begin{figure}[t]
    \centering
    \includegraphics[width=\textwidth]{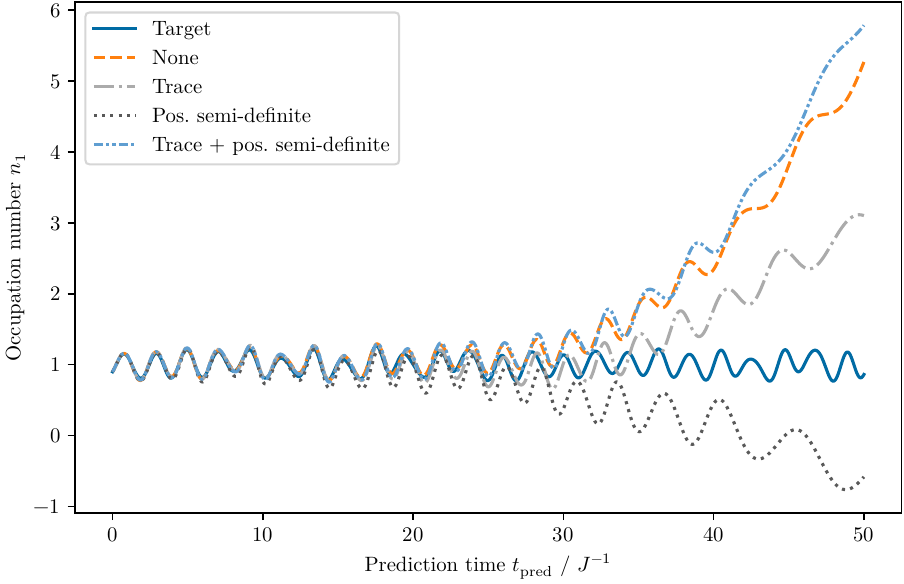}
    \caption{The occupation number $n_1$ of site 1 of the six-site Fermi-Hubbard model described in Eq.~\eqref{eq:fermi-hubbard}, with $V=1.0$ and $U=3.1$, is visualized over time by the blue solid line. The occupation number is calculated from the \ac{2rdm}. The predicted occupation number of four hyper-parameter optimized models trained on different losses are shown as well, all trained on the first 3000 time steps ($30\,J^{-1}$) of the time series. The orange dashed line was trained on the MSE loss only, the light gray dash-dotted line additionally on the constraint loss for the trace $\mathcal{L}_\text{tr}$, and the dark gray dotted line on the MSE and the negative eigenvalue loss $\mathcal{L}_\text{psd}$. The prediction starts from $D_{12} (t=40\,J^{-1})$, hence this is an extrapolation.}
    \label{fig:occupation-constraints}
\end{figure}

In the previous section, we focused on the short- to medium-term forecast. In Fig.~\ref{fig:occupation-constraints}, we consider the long-term prediction. The prediction diverges after a prediction time of about $t_\text{pred} = 25 - 30 \, J^{-1}$. We note that Van de Walle et al.~\cite{vandewalle2024ManyBody}, who report state-of-the-art extrapolation capabilities, demonstrate predictions only up to $t_\text{pred} = 8 \, J^{-1}$, with considerably fewer oscillations than those encountered in Fig.~\ref{fig:occupation-constraints}.
Interestingly, the same behavior is observed in the analytical \ac{td2rdm} theory \cite{donsa2023Nonequilibrium}. To stabilize the \ac{td2rdm} method, several necessary $N$-representability conditions are enforced, if broken, during integration by means of a purification procedure \cite{donsa2023Nonequilibrium,garrod1964Reduction,coleman1963Structure,mazziotti2012Structure,pescoller_projective_2025}. Specifically, the positive semi-definiteness of the \ac{2rdm} and the \ac{2hrdm} are imposed. The matrix elements of the 2HRDM $Q_{12}(t)$ are given by $Q^{j_1\uparrow j_2\downarrow}_{i_1\uparrow i_2\downarrow} = \langle \Psi(t)|a_{j_1\uparrow} a_{j_2\downarrow} a_{i_1\uparrow}^\dagger a_{i_2\downarrow}^\dagger|\Psi(t) \rangle$. Empirically, it has been found that this purification procedure stabilizes the \ac{td2rdm} very well in all parameter regions, where the time-local reconstruction functionals produce a sufficiently small error in the overall propagation \cite{pescoller_projective_2025}.
A direct benchmark between neural ODE and \ac{td2rdm} for this representative setting is available in the public data repository associated with this study~\cite{DARUS}.

The question arises whether we can enforce these conditions on the neural \ac{ode} as well? Do they stabilize the model's long-term predictions? How do they perform compared to the unconstrained model? To enforce constraints in \ac{ml}, there are, in principle, two options: i) weak constraints enforce the conditions via the loss function during training, while ii) hard constraints are built into the architecture of the model itself, preventing the model from violating these constraints. In the following, we will focus on weak constraints and their results.

\subsubsection{Weak constraints}

The week constraints are enforced via the loss function
\begin{align}
    \mathcal{L} = \text{MSE}(\hat{y}, y) + \sum_i \alpha_i \mathcal{L}_{\text{constraint}, i}\,,
\end{align}
with the mean square error as standard loss and the different constraints $\mathcal{L}_{\text{constraint}, i}$ scaled with a factor $\alpha_i$.

Due to their positive semi-definiteness, the negative eigenvalues $\lambda_j^t$ for both the \ac{2rdm} $D_{12}(t)$ and the \ac{2hrdm} $Q_{12}(t)$ at time $t$ are
penalized via
\begin{align}
    \mathcal{L}_\text{psd} &= \frac{1}{T} \sum_{t=0}^T \sum_j \text{ReLU}\left( - \lambda_j^t \right)^2\,.
    \label{eq:constraint_eigval}
\end{align}
The rectified linear unit ($\text{ReLU}$) is squared to obtain a continuous derivative.
\begin{figure}[tp]
    \centering
    \includegraphics[width=\textwidth]{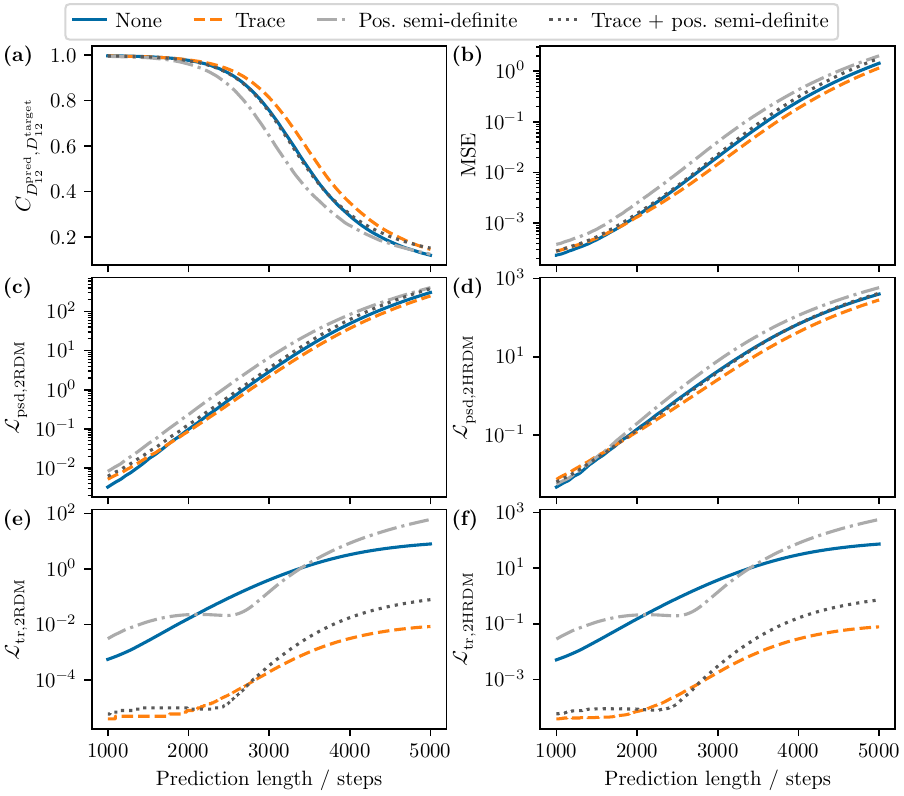}
    \caption{The Pearson correlation coefficient $C_{D_{12}^\text{pred},D_{12}^\text{target}}$ (a), the MSE loss (b), the eigenvalue loss for the \ac{2rdm} $\mathcal{L}_{\text{psd,2RDM}}$ (c), and the \ac{2hrdm} $\mathcal{L}_{\text{psd,2HRDM}}$ (d), as well as the trace loss for the \ac{2rdm} $\mathcal{L}_{\text{tr,2RDM}}$ (e) and the \ac{2hrdm} $\mathcal{L}_{\text{tr,2HRDM}}$ (f) for models trained on different losses are all visualized over the length of the prediction. The blue solid line represents the unconstrained model, which was trained only on MSE loss. Models that had either trace constraint or negative eigenvalue constraint as additional loss terms are depicted as an orange dashed line and a light gray dashed-dotted line, respectively. The dark gray dotted line indicates the model trained with all of the previously mentioned loss terms.}
    \label{fig:performance_prediction-length}
\end{figure}
Thus, we have two additional terms in the loss, one for the \ac{2rdm} and one for the \ac{2hrdm}.
Similarly, the trace conservation of the \ac{2rdm} over time
\begin{align}
    \sum_n\sum_m D^{n\uparrow m\downarrow}_{n\uparrow m\downarrow} (t) = N_\uparrow N_\downarrow = 9, ~ \forall ~ t\,
\end{align}
is enforced through the loss
\begin{align}
    \mathcal{L}_\text{tr} &= \frac{1}{T} \sum_{t=0}^T \left( \sum_n\sum_m D^{n\uparrow m\downarrow}_{n\uparrow m\downarrow} (t) - 9 \right)^2
    \label{eq:constraint_norm}
\end{align}
for the \ac{2rdm} $D_{12}(t)$ and analogously for the \ac{2hrdm} $Q_{12}(t)$.

For the following analysis, we focus on the parameter configuration $V=1.0\,J$ and $U=3.1\,J$ because of its dynamics, which is neither too slow nor too fast. Additionally, it is in the correlated regime, where the prediction should, in principle, work, but the performance in Fig.~\ref{fig:correlations_performance}(d) is not optimized ($C_{D_{12}^\text{pred},D_{12}^\text{target}} = 0.6899$).
First, we optimize the model's hyperparameters, resulting in much better performance: $C_{D_{12}^\text{pred},D_{12}^\text{target}} = 0.9194$. This supports the previous statement that the fluctuations in the performance in Fig.~\ref{fig:correlations_performance}(c) and (d) could be reduced by hyper-parameter optimization. Next, we train three models on different combinations of the constraints, one on the trace constraint only, one on the eigenvalue constraint only, and one on both. Again, we optimize the hyperparameters of each model for a fair comparison. As depicted in Fig.~\ref{fig:occupation-constraints}, these constraints do not stabilize the long-term prediction. In fact, they all start to diverge at a similar prediction length, as defined in Eq.~\eqref{eq:prediction-length}, $25\,J^{-1} < t_\text{pred} < 30\,J^{-1}$. In Fig.~\ref{fig:performance_prediction-length} the Pearson correlation coefficient $C_{D_{12}^\text{pred},D_{12}^\text{target}}$ and the different loss terms over the length of the prediction are visualized.
The same divergent behavior is observed in the Pearson correlation. All models show similar performance across the various loss metrics, except for trace loss, where models trained with this constraint perform about two orders of magnitude better. However, this only leads to a minimally longer stable forecast length in the case of the trace constraint model, while there is no improvement in the Pearson correlation coefficient (see Fig.~\ref{fig:performance_prediction-length}(a)) for the model trained with both constraints. In general, the losses for the \ac{2rdm} are lower than those for the \ac{2hrdm}, which is to be expected since the model is only trained directly with the \ac{2rdm} data, while the \ac{2hrdm} is derived from it. Interestingly, for short-term predictions up to a prediction length of $t_\text{pred} = 1700\,\text{steps} = 17\,J^{-1}$ the unconstrained model performs best, but only slightly. This can be seen in the loss plots of the MSE and the eigenvalue constraint.

We note that training the constraint models took significantly longer, as the eigenvalues and trace had to be calculated for each time step of the prediction.

\section{Discussion and Conclusion}
\label{sec:discussion}

Non-equilibrium dynamics in the Fermi-Hubbard model is Markovian on the level of the full many-body Schr\"{o}dinger equation. Intricate non-Markovian behavior arises upon reduction of degrees of freedom, e.g.~see Ref.~\cite{Coppola2025Learning}. Tracing out degrees of freedom from the full wave function leads to reduced density matrices such as the \ac{2rdm}. In this work, we showed how the neural \ac{ode} serves as a model‑agnostic diagnostic tool to identify the degree of Markovianity of the \ac{2rdm} $D_{12}(t)$ dynamics. Its ability to reproduce the exact trajectories without any explicit knowledge about the \ac{3rdm} confirms that, in regimes where the Pearson coefficient between the two- and three-particle cumulants $C_{\Delta_{12},\Delta_{123,\text{K}}}$ is large, the dynamics is effectively governed by a time‑local functional of the instantaneous two‑particle cumulant $\Delta_{12}(t)$. This means that no memory is required for the propagation. In the anti‑correlated regime, where $C_{\Delta_{12},\Delta_{123,\text{K}}}$ becomes negative, the neural \ac{ode} fails to predict the evolution, indicating that no simple time‑local reconstruction of the three‑particle cumulant $\Delta_{123}(t)$ exists. In other words, memory is required.

We want to emphasize that the neural \ac{ode} should not replace exact diagonalization or other physical algorithms, but rather tests whether the dynamics can be treated as Markovian. Since Markovianity is a widely used assumption in many methods and theories across different fields, we argue that a dedicated tool to verify this assumption is valuable, even at the computational cost observed here.
At the same time, the proposed framework is not restricted to quantum systems.
It can be applied to other types of time-series data, including settings with substantially lower dimensionality, where smaller models are sufficient and computational cost can be reduced significantly.
Therefore, if exact reference data are available, we can ask a concrete and practically relevant question before committing to a given approximation scheme: is this specific assumption (Markovianity) justified for the dynamics at hand?
As discussed in the introduction, existing Markovianity tests are either restricted to specific settings (e.g., one-dimensional reaction coordinates~\cite{Berezhkovskii2018SingleMolecule}, open-system quantum channels~\cite{wolf2008Assessing}, or stationary stochastic processes~\cite{Chen2012Testing,zhou2023Testing,shi2020Does}) or provide only a binary accept/reject verdict. The neural \ac{ode} framework fills this gap by offering a general, assumption-free diagnostic that yields a continuous measure of effective Markovianity and is directly applicable to any temporal coherent time-series data, including the high-dimensional, deterministic, non-stationary dynamics of reduced density matrices.

Improving the quality of the extrapolated dynamics via neural \acp{ode} benefits remarkably strongly from exploring the hyper-parameter space to find the near-optimal parameters. Hyper-parameter optimization works even better to obtain robust results than model-ensemble averaging. We achieve an extrapolation of several thousand time-steps, corresponding to a prediction time between $25 \, J^{-1} < t_\text{pred} < 30 \, J^{-1}$ in our model.

An analysis of the buildup of genuine three‑particle correlations $\overline{\delta\Delta^{\uparrow\uparrow\downarrow}_{123,\text{K}}}$ reveals that the magnitude of $\overline{\delta\Delta^{\uparrow\uparrow\downarrow}_{123,\text{K}}}$ is the primary factor limiting the applicability of current reconstruction functionals. Regions with $\overline{\delta\Delta^{\uparrow\uparrow\downarrow}_{123,\text{K}}} < 0.65$ coincide with accurate neural \ac{ode} forecasts and low reconstruction errors, whereas larger values correlate with the breakdown of both the neural \ac{ode} and the \ac{td2rdm} method.

These observations suggest a natural extension of the neural \ac{ode} framework: by endowing the vector field with explicit history dependence -- e.g., with an explicit history channel, where the vector field depends on both the current state and a learned memory state, $\dot{y}(t)=f_\theta\big(y(t),m(t)\big)$ with $\dot{m}(t)=g_\phi\big(m(t),y(t)\big)$, or equivalently a finite-history formulation $\dot{y}(t) = f_\theta \big(y(t), y(t-\tau_1), \ldots, y(t-\tau_k)\big)$ -- one can empirically determine the minimal memory length required for an accurate reconstruction of $\Delta_{123}(t)$. Such a memory‑aware neural \ac{ode} would provide quantitative guidance for developing non‑local reconstruction functionals that incorporate the necessary temporal kernels.

During training, we imposed weak physical constraints (norm conservation and positive semi-definiteness) via additional loss terms. Although the norm constraint improves predictions slightly, the constraints did not yield a noticeable enhancement of long‑time stability. Enforcing hard constraints is more challenging for neural \acp{ode} because they learn derivatives, thus symmetries that apply on the level of the integrated observable cannot be implemented straightforwardly into the architecture of the model.
Recent proposals for constrained neural differential equations as in Ref.~\cite{white2023Stabilized} offer promising routes to incorporate such hard constraints.

Finally, this study shows that neural ODEs can be trained on high-dimensional data even with few data samples (with dimensionality of the same order as the number of samples in the training set), which, to the authors' knowledge, has not yet been described in literature. Despite the large state space, the learned model provides reliable short‑ to medium‑term predictions (up to $t_\text{pred}\approx30\,J^{-1}$ = 3000\,\text{steps}) across a broad range of interaction strengths $U$ and quench amplitudes $V$, which opens up new application possibilities.

\section*{Acknowledgments}
We thank Katharina Buczolich, Elias Pescoller, Marie Eder, Florian Grüner, Mathias Niepert, Maria Wirzberger, Max Weinmann, and Mario Gaimann for helpful and stimulating discussions.
MK and PE are gratefully funded by Deutsche Forschungsgemeinschaft (DFG, German Research Foundation) under Germany’s Excellence Strategy -- EXC 2075 -- 390740016 and PE by the Ministry of Science, Research and the Arts Baden-Württemberg (Az. 33-7533-9-19/54/5) in ``Künstliche Intelligenz \& Gesellschaft: Reflecting Intelligent Systems for Diversity, Demography and Democracy'' (IRIS3D).
We thank the Stuttgart Center for Simulation Science (SimTech) and the Interchange Forum for Reflecting on Intelligent Systems (IRIS) at the University of Stuttgart, the Heidelberg Academy of Sciences and Humanities, and (PE) the International Max Planck Research School for Intelligent Systems (IMPRS-IS) for further support.
IB thanks the Austrian Research Fund (FWF) through grant P35539-N for support.
Computations have been performed on the Austrian Scientific Cluster (ASC) and the Experimental Compute Cluster: ehlers at the University of Stuttgart.

\section*{Data availability}
The data that support the findings of this study are openly available at the following URL/DOI: \url{https://
doi.org/10.18419/DARUS-5613}~\cite{DARUS}.

\printbibliography

\end{document}